\documentclass[sigconf]{acmart}

\AtBeginDocument{%
  }

\copyrightyear{2026}
\acmYear{2026}
\setcopyright{cc}
\setcctype{by}
\acmConference[SIGIR '26] {Proceedings of the 49th International ACM SIGIR Conference on Research and Development in Information Retrieval}{July 20--24, 2026}{Melbourne, VIC, Australia.}
\acmBooktitle{Proceedings of the 49th International ACM SIGIR Conference on Research and Development in Information Retrieval (SIGIR '26), July 20--24, 2026, Melbourne, VIC, Australia}
\acmISBN{979-8-4007-2599-9/2026/07}
\acmDOI{10.1145/3805712.3808555}

\usepackage{booktabs, multirow, makecell}
\usepackage{siunitx}
\usepackage{array}
\usepackage{xspace}

\usepackage{graphicx}
\usepackage{xcolor, colortbl}
\usepackage{array}
\usepackage{enumitem}

\usepackage[most]{tcolorbox}

\newtcolorbox{promptbox}[1]{
    colback=blue!5!white,
    colframe=blue!75!black,
    fonttitle=\bfseries,
    title=#1,
    arc=0mm,
    boxrule=0.5pt
}

\newcommand{\rqheading}[1]{\noindent\textbf{#1}\xspace}

\begin{document}
\title{Reproducing Complex Set-Compositional Information Retrieval}

\author{Vincent Degenhart}
\affiliation{%
  \institution{Radboud University}
  \city{Nijmegen}
  \country{NL}
}
\email{vincent.degenhart@ru.nl}
\orcid{0009-0009-8823-0022}

\author{Dewi Timman}
\affiliation{%
  \institution{Radboud University}
  \city{Nijmegen}
  \country{NL}
}
\email{dewi.timman@ru.nl}
\orcid{0009-0007-7831-4359}

\author{Arjen P.\ de Vries}
\affiliation{%
  \institution{Radboud University}
  \city{Nijmegen}
  \country{NL}
}
\email{arjen.devries@ru.nl}
\orcid{0000-0002-2888-4202}

\author{Faegheh Hasibi}
\affiliation{%
  \institution{Radboud University}
  \city{Nijmegen}
  \country{NL}
}
\email{faegheh.hasibi@ru.nl}
\orcid{0009-0006-9986-482X}

\author{Mohanna Hoveyda}
\affiliation{%
  \institution{Radboud University}
  \city{Nijmegen}
  \country{NL}
}
\email{mohanna.hoveyda@ru.nl}
\orcid{0009-0003-8027-6575}

\begin{abstract}
Complex information needs may involve set-compositional queries using conjunction, disjunction, and exclusion, yet it remains unclear whether current retrieval paradigms genuinely satisfy such constraints or exploit `semantic shortcuts'. We conduct a reproducibility study to benchmark major retrieval families and reasoning-targeted methods on QUEST and QUEST+Variants, and introduce LIMIT+, a controlled benchmark where relevance depends on arbitrary attribute predicates and constraint satisfaction, and less on pretrained knowledge. Our findings show that (i) on QUEST, the best neural retrievers achieve an effectiveness that is more than double what can be achieved with BM25 (Recall@100 ${>}$0.41 vs.\ 0.20), but reasoning-targeted methods like ReasonIR and Search-R1 do not outperform general-purpose retrievers uniformly; (ii) on LIMIT+, gains fail to transfer, where the strongest QUEST method collapses from Recall@100${\approx}$0.42 to below 0.02, while classic lexical retrieval gains to ${\sim}$0.96. Lastly, (iii) stratifying by compositional depth reveals a consistent degradation across all methods, where algebraic sparse and lexical methods show more stable performance while dense approaches collapse.
We release code and LIMIT+ data generation scripts to support future reproducibility and controlled evaluation.
\end{abstract}



\maketitle

\section{Introduction}

In many real-world deployments, information needs involve more than topical or semantic similarity, and need to satisfy several conditions, avoiding results that violate these constraints~\cite{weller-etal-2024-nevir}. A user may seek entities that match several properties, or match one property while excluding another~\cite{zhang2025excluir, malaviya-etal-2023-quest}. These information needs can be formalized as set operations; intersection (AND), union (OR), and set difference (NOT). 

\begin{figure}[th]
    \centering
    \includegraphics[width=0.95\columnwidth]{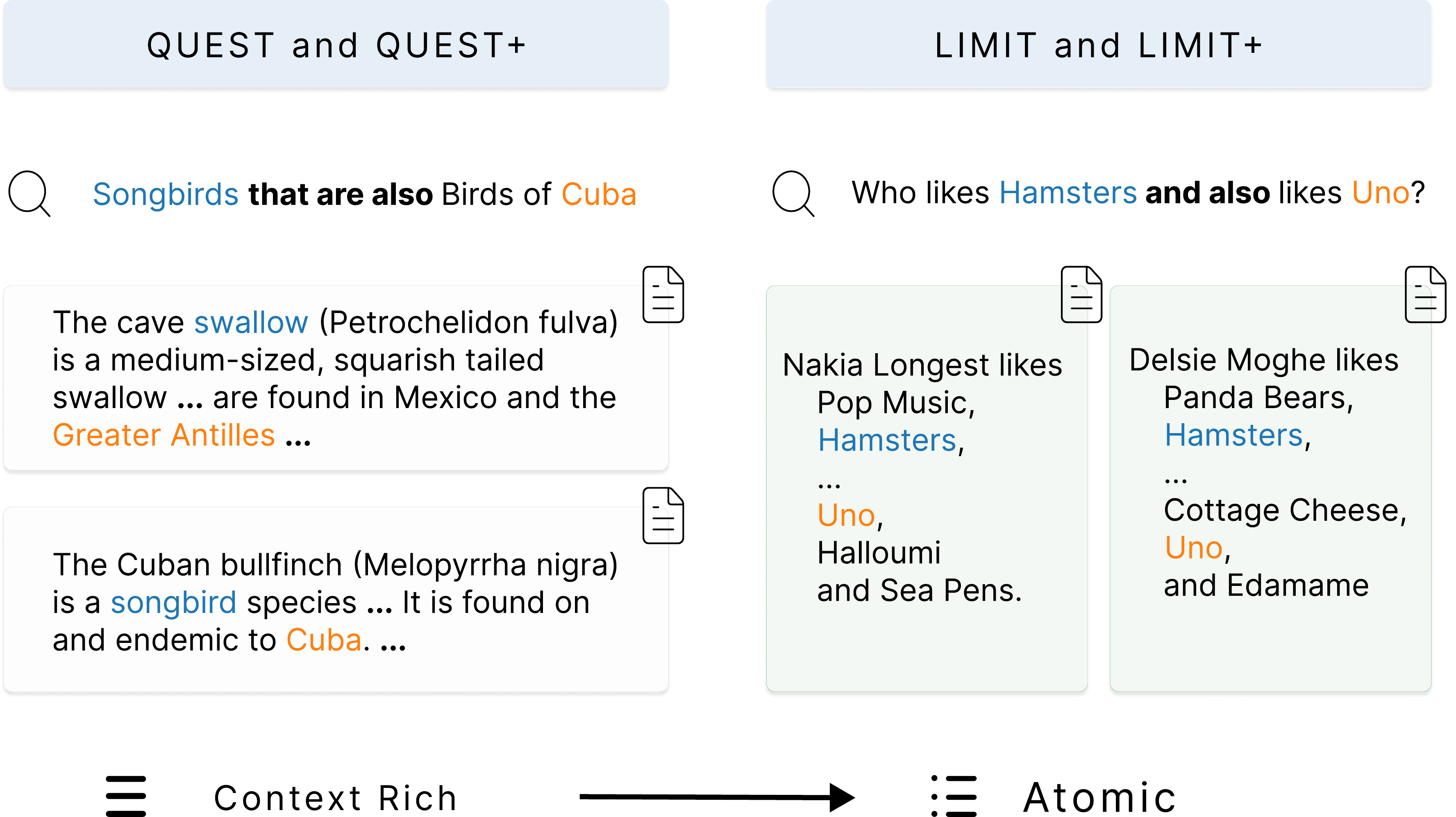}
    \phantomcaption
    \caption{The transition from context-rich narrative to atomic datasets.}
    \label{fig:placeholder}
    \Description[Datasets example.]{Figure to illustrate the datasets used.}
\end{figure}
The urgency of addressing these operations is underscored by two trends:

\begin{description}    
    \item[Deep Research:] The frequency of constraint-heavy queries increases as users engage in more specialized, deep-research tasks that move beyond simple keyword matching \cite{zheng-etal-2025-deepresearcher}.

    \item[Agentic Query Rewriting:] In agentic workflows, autonomous systems frequently rewrite user prompts into multi-step logical queries, requiring the underlying retrieval engine to be more than just semantically aware \cite{wu-etal-2025-agentic}.
\end{description}

Yet most modern retrieval systems are trained and evaluated on benchmarks where semantic relatedness is a dominant signal ~\cite{Bajaj2016-uj, Craswell2021-kx}. Consequently, improvements on complex-query retrieval benchmarks can be difficult to interpret: strong aggregate performance does not, by itself, demonstrate reliable enforcement of compositional requirements, and may instead reflect semantic heuristics and pretrained world knowledge that correlate with the query constraints. Recent benchmarks such as \textsc{QUEST}~\cite{malaviya-etal-2023-quest, shen-etal-2025-logicol} operationalize set-compositional retrieval by instantiating entity-seeking queries whose templates imply set operations over a large Wikipedia corpus. In parallel, retrieval research now spans a broad range of paradigms, from lexical and dense bi-encoders to late-interaction, learned sparse retrieval, and LLM-based rerankers, alongside recent methods proposed specifically for reasoning-intensive retrieval via LLMs or enhanced constraint representation learning. This motivates two empirical questions. Firstly, it is unclear how major retrieval families compare under a consistent, reproducible evaluation protocol in complex set-compositional retrieval. Second, even when aggregate scores improve, it is unclear whether observed gains reflect improved adherence to compositional requirements or improvements driven by correlations in the underlying data and model priors.

To address these questions, we conduct a reproducibility study with a two-part evaluation.
We first benchmark a broad suite of retrievers and rerankers on \textsc{QUEST} and \textsc{QUEST+Variants}~\cite{shen-etal-2025-logicol}
under a unified pipeline that standardizes indexing, retrieval, and evaluation. We then introduce \textsc{LIMIT+}, a new benchmark designed to isolate constraint satisfaction from semantic shortcut effects. \textsc{LIMIT+} extends \textsc{LIMIT}~\cite{weller2025theoreticallimitationsembeddingbasedretrieval} by composing atomic attribute predicates into \textsc{QUEST}-style templates, so that relevance is determined by exact predicate satisfaction rather than topical relatedness. This construction yields exact gold sets by design and associates each query instance with template and attribute metadata, enabling stratified analyses by constraint type and compositional depth.
\subsubsection*{Research Questions} Concretely, we study the following:
\begin{enumerate}[leftmargin=*, label={}, itemsep=2pt, topsep=2pt]
  \item \textbf{RQ1.} How do major retrieval families and reasoning-targeted methods (e.g., Search-R1~\cite{jin2025search}, ReasonIR~\cite{shao2025reasonir}) perform on set-compo-
  sitional queries in \textsc{QUEST} and \textsc{QUEST+Variants}?
  \item \textbf{RQ2.} Do gains observed on \textsc{QUEST} persist in a controlled setting  in LIMIT+ (Section \ref{sec:limit+}) where semantic similarity provides limited signal and relevance is driven by constraint satisfaction?
  \item \textbf{RQ3.} How does effectiveness change with increasing compositional depth and across constraint types?
\end{enumerate}

\subsubsection*{Summary of main findings}
Our experiments reveal 3 key results.

\emph{(RQ1)} On \textsc{QUEST} and \textsc{QUEST+Variants}, retrieval families separate clearly: the best neural retrievers demonstrate a twofold improvement over BM25 (Recall@100 ${>}$0.41 vs.\ 0.20), yet methods explicitly designed for reasoning in retrieval, such as ReasonIR and Search-R1, do not outperform strong general-purpose retrievers; suggesting that a reasoning specific objective is not sufficient to perform well on set compositional retrieval tasks.
\emph{(RQ2)} When semantic shortcuts are minimized in \textsc{LIMIT+}, the performance gains observed on simpler benchmarks fail to generalize: the strongest enhanced method on \textsc{QUEST} (LogiCol; Recall@100${\approx}$0.42) collapses below 0.02, and all dense single-vector models fall below 0.10, while lexical retrieval (${\approx}$0.96) and token-level methods remain robust, suggesting that much of the observed progress on \textsc{QUEST} reflects learned correlations with semantic cues rather than genuine constraint satisfaction. 
Under a controlled re-ranking setup where all gold documents are guaranteed to appear in the candidate set, the LLM-based re-ranker achieves near-ceiling effectiveness (nDCG@5 up to 0.97). The neuro-symbolic approach is a close second, trailing by only a marginal gap, while the cross-encoder fails at successfully positioning relevant documents on the top (${\approx}$0.16). This localizes the primary bottleneck in set-compositional complex queries to first-stage candidate generation.
\emph{(RQ3)} Lastly, stratifying by constraint type and compositional depth reveals that aggregate scores mask heterogeneity: templates with negations, in isolation or combined with other constraints, are markedly harder than atomic or disjunctive queries, and performance degrades monotonically with depth. On \textsc{LIMIT+}, algebraic sparse methods (\cite{krasakis2025constructing}) remain stable (nDCG@20${\approx}$0.22) where learned dense approaches collapse (${\approx}$0.01).  

\subsubsection*{Contributions}

\begin{enumerate}
    \item A reproducibility study benchmarking lexical, dense, late-interaction, learned sparse, cross-encoder, LLM and neuro-symbolic reranking pipelines, alongside reasoning-oriented methods (ReasonIR, Search-R1, LogiCol~\cite{shen-etal-2025-logicol}, Set-Comp~\cite{krasakis2025constructing}), under a unified evaluation protocol for set-compositional retrieval.
    \item  \textsc{LIMIT+}, a controlled set-compositional benchmark derived from \textsc{LIMIT} by composing \textsc{QUEST}-style constraint templates with controlled depth and constraint types, enabling stratified analyses that disentangle semantic relatedness from constraint satisfaction.
    \item Diagnostic ablations over compositional depth and constraint types, exposing failure modes invisible in aggregate metrics, showing how learned dense representations and some reasoning-enhanced methods collapse under complex composition.
\end{enumerate}

\noindent The remainder of the paper is organized as follows. Section~\ref{sec:background} surveys related work on complex-query retrieval, retrieval paradigms, and reasoning-targeted methods. Section~\ref{sec:limit+} describes the construction of the \textsc{LIMIT+} benchmark. Section~\ref{sec:exp} details the experimental setup, including the models, indexing pipelines, and evaluation protocol used to ensure reproducibility. Section~\ref{sec:res} presents results on \textsc{QUEST}, \textsc{QUEST+Variants}, and \textsc{LIMIT+}, together with constraint-type and depth-stratified analyses. Section~\ref{sec:conc} discusses implications and concludes. Additional experimental details, prompts, and a replication comparison against originally reported scores are provided in the Appendix.

\sisetup{
  detect-weight=true,
  detect-family=true,
  table-number-alignment = center,
  round-mode=places,
  round-precision=3,
  table-format=1.3
}

\setlength{\tabcolsep}{4pt}
\newcommand{\dsGutter}{6pt}     %

\newcommand{\n}[1]{\num{#1}}

\newcommand{\kone}{5}
\newcommand{\ktwo}{20}
\newcommand{\kthree}{100}
\newcommand{\ktriple}{%
  \makecell{\scriptsize @\kone} &
  \makecell{\scriptsize @\ktwo} &
  \makecell{\scriptsize @\kthree}%
}

\begin{table*}[t]
\centering
\small
\renewcommand{\arraystretch}{1.08}
\caption{Reproducibility results on complex-query retrieval datasets with set-compositional queries (QUEST). Metrics show Recall and nDCG @5,20,100 for first-stage retrieval. Params and training objectives are summarized for each model. Scores in bold indicate the highest score in that column.}
\label{tab:repro_results_quest}
\begin{tabular}{
  l c >{\raggedright\arraybackslash}p{2.3cm}
   @{}r r r r r r @{\hspace{\dsGutter}} %
    r r r r r r                          %
}
\toprule
\multicolumn{3}{c}{} &
\multicolumn{6}{c}{\bf QUEST} &
\multicolumn{6}{c}{\bf QUEST+Variants} \\
\cmidrule(lr){4-9}\cmidrule(lr){10-15}
\multicolumn{3}{c}{} &
\multicolumn{3}{c}{\bf Recall} & \multicolumn{3}{c}{\bf nDCG} &
\multicolumn{3}{c}{\bf Recall} & \multicolumn{3}{c}{\bf nDCG} \\
\cmidrule(lr){4-6}\cmidrule(lr){7-9}
\cmidrule(lr){10-12}\cmidrule(lr){13-15}
\textbf{Model} & \textbf{Params} & \textbf{Train.\ obj.} &
\ktriple & \ktriple & \ktriple & \ktriple \\
\midrule

\rowcolor[HTML]{F2F2F2}
\multicolumn{15}{l}{\bf Lexical} \\
BM25 \cite{robertson1994okapi} & -- & -- &
\n{0.047} & \n{0.104} & \n{0.1967} & \n{0.0949} & \n{0.0942} & \n{0.130} &
\n{0.0188} & \n{0.0432} & \n{0.0932} & \n{0.0564} & \n{0.0544} & \n{0.0637} \\
\midrule
\rowcolor[HTML]{F2F2F2}
\multicolumn{15}{l}{\bf Dense (single-vector)} \\
Contriever \cite{izacard2021contriever} & 110M & InfoNCE &
\n{.0362} & \n{.0787} & \n{.1461} & \n{.0522} & \n{.0638} & \n{.0866} &
\n{0.0139} & \n{0.0319} & \n{0.0733} & \n{0.0497} & \n{0.0477} & \n{0.0531} \\
Promptriever \cite{weller2024promptriever} & 7B & InfoNCE &
\n{0.0271} & \n{0.0514} & \n{0.0920} & \n{0.0680} & \n{0.0546} & \n{0.0703} &
\n{0.0146} & \n{0.0277} & \n{0.0533} & \n{0.0565} & \n{0.0470} & \n{0.0452} \\
Qwen3-Embedding (8B) \cite{qwen3embedding} & 8B & InfoNCE &
\n{0.0950} & \n{0.2027} & \n{0.3839} & \n{0.1923} & \n{0.1862} & \n{0.2574} &
\n{0.0381} & \n{0.0903} & \n{0.1958} & \n{0.1182} & \n{0.1193} & \n{0.1405} \\

E5-Mistral-Instruct \cite{WangLiang2023ITEw} & 7B & InfoNCE &
\n{0.0853} & \n{0.1787} & \n{0.3358} & \n{0.1777} & \n{0.1669} & \n{0.2288} &
\n{0.0357} & \n{0.0799} & \n{0.1695} & \n{0.1074} & \n{0.1067} & \n{0.1235} \\
E5-base-V2 \cite{WangLiang2022TEbW} & 110M & InfoNCE &
\n{0.0794} & \n{0.1663} & \n{0.3181} & \n{0.1608} & \n{0.1517} & \n{0.2098} &
\n{0.0329} & \n{0.0726} & \n{0.1580} & \n{0.1064} & \n{0.1026} & \n{0.1162} \\
GritLM-7B \cite{muennighoff2024generative} & 7B & LM+contrast &
\n{0.1067} & \n{0.2253} & \bf\n{0.4193} & \bf\n{0.2181} & \n{0.2091} & \bf\n{0.2844} &
\bf\n{0.0465} & \n{0.1028} & \n{0.2111} & \n{0.1388} & \n{0.1363} & \n{0.1564} \\
\midrule
\rowcolor[HTML]{F2F2F2}
\multicolumn{15}{l}{\bf Dense (multi-vector)} \\
GTE-ModernColBERT \cite{GTE-ModernColBERT} & 0.1B & KL-Div+CE &
\num{0.0999} & \num{0.1890} & \num{0.2944} & \num{0.1984} & \num{0.1839} & \num{0.2286} & \num{0.0403} & \num{0.0865} & \num{0.1500} & \num{0.1208} & \num{0.1166} & \num{0.1201} \\
\midrule

\rowcolor[HTML]{F2F2F2}
\multicolumn{15}{l}{\bf Learned Sparse Retrieval (LSR)} \\
SPLADE \cite{lassance2024spladev3} & 110M & KL-Div+MSE &
\n{0.1019} & \n{0.2117} & \n{0.3927} & \n{0.1922} & \n{0.1910} & \n{0.2621} &
\n{0.0387} & \n{0.0871} & \n{0.1863} & \n{0.1077} & \n{0.1101} & \n{0.1331} \\
\midrule

\rowcolor[HTML]{F2F2F2}
\multicolumn{15}{l}{\bf Enhanced Methods} \\
Search-R1 (with BM25) \cite{jin2025search} & 7B & EM+PPO &
\n{0.0529} & \n{0.1159} & \n{0.2279} & \n{0.1013} & \n{0.1033} & \n{0.1471} &
\num{0.0220} & \num{0.0512} & \num{0.1101} & \num{0.0641} & \num{0.0643} & \num{0.0760} \\

Set-Comp LSR \cite{krasakis2025constructing} & 0 & Zeroshot &
\n{0.0907} & \n{0.1775} & \n{0.2888} & \n{0.1795} & \n{0.1705} & \n{0.2170} &
\n{0.0407} & \n{0.0879} & \n{0.1797} & \n{0.1167} & \n{0.1171} & \n{0.1347} \\

ReasonIR-8B \cite{shao2025reasonir} & 8B & InfoNCE &
\n{0.0819} & \n{0.1843} & \n{0.3636} & \n{0.1697} & \n{0.1669} & \n{0.2373} &
\n{0.0386} & \n{0.0898} & \n{0.1919} & \n{0.1196} & \n{0.1202} & \n{0.1395} \\
LogiCol \cite{shen-etal-2025-logicol} & 110M & Mod. contrastive &
\textbf{\n{0.1145}} & \textbf{\n{0.2342}} & \n{0.4188} & \n{0.2157} & \bf\n{0.2128} & \bf\n{0.2844} &
\bf\n{0.0465} & \bf\n{0.1032} & \bf\n{0.2168} & \bf\n{0.1458} & \bf\n{0.1436} & \bf\n{0.1627} \\
\bottomrule
\end{tabular}
\medskip
\\
\footnotesize
\textit{Train.\ obj.\ abbreviations:} InfoNCE = contrastive loss; KL-Div+CE = KL-divergence + cross-entropy; KL-Div+MSE = KL-divergence + MarginMSE; LM+contrast = language modeling + contrastive embedding; EM+PPO = exact match reward + PPO; Mod.\ contrastive = modified supervised contrastive loss.
\end{table*}

\sisetup{
  detect-weight=true,
  detect-family=true,
  table-number-alignment = center,
  round-mode=places,
  round-precision=3,
  table-format=1.3
}

\setlength{\tabcolsep}{3pt}     %

\begin{table*}[t]
\centering
\small
\renewcommand{\arraystretch}{1.08}
\caption{Reproducibility results on complex-query retrieval datasets with set-compositional queries (LIMIT). Metrics show Recall and nDCG @5,20,100. Scores in bold indicate the highest score in that column.}
\begin{tabular}{
  l l
   @{}r r r r r r @{\hspace{\dsGutter}} %
    r r r r r r                          %
}
\toprule
\multicolumn{2}{c}{} &
\multicolumn{6}{c}{\bf LIMIT} &
\multicolumn{6}{c}{\bf LIMIT+Constraints} \\
\cmidrule(lr){3-8}\cmidrule(lr){9-14}
\multicolumn{2}{c}{} &
\multicolumn{3}{c}{\bf Recall} & \multicolumn{3}{c}{\bf nDCG} &
\multicolumn{3}{c}{\bf Recall} & \multicolumn{3}{c}{\bf nDCG} \\
\cmidrule(lr){3-5}\cmidrule(lr){6-8}
\cmidrule(lr){9-11}\cmidrule(lr){12-14}
\multicolumn{2}{c}{} &
\ktriple & \ktriple & \ktriple & \ktriple \\
\midrule

\rowcolor[HTML]{F2F2F2}
\multicolumn{14}{l}{\bf Lexical} \\
BM25 \cite{robertson1994okapi} & &
\bf\num{0.9440} & \bf\num{0.9535} & \bf\num{0.9635} & \bf\num{0.9392} & \bf\num{0.9427} & \bf\num{0.9449} & \bf\num{0.6321} & \bf\num{0.7325} & \bf\num{0.8372} & \bf\num{0.7577} & \bf\num{0.7853} & \bf\num{0.8013} \\
\midrule

\rowcolor[HTML]{F2F2F2}
\multicolumn{14}{l}{\bf Dense (single-vector)} \\
Contriever \cite{izacard2021contriever}& &
\num{0.0325} & \num{0.0665} & \num{0.1460} & \num{0.0270} & \num{0.0387} & \num{0.0562} & \num{0.0215} & \num{0.0495} & \num{0.1115} & \num{0.0796} & \num{0.0748} & \num{0.0751} \\
Promptriever \cite{weller2024promptriever} & &
\num{0.0355} & \num{0.0405} & \num{0.0475} & \num{0.0373} & \num{0.0390} & \num{0.0406} &
\num{0.0159} & \num{0.0198} & \num{0.0340} & \num{0.0273} & \num{0.0257} & \num{0.0258} \\
Qwen3-Embedding (8B) \cite{qwen3embedding} & &
\num{0.0150} & \num{0.0240} & \num{0.0505} & \num{0.0123} & \num{0.0154} & \num{0.0212} &
\num{0.0125} & \num{0.0199} & \num{0.0446} & \num{0.0273} & \num{0.0278} & \num{0.0299} \\
E5-Mistral-Instruct \cite{WangLiang2023ITEw} & &
\num{0.0155} & \num{0.0265} & \num{0.0600} & \num{0.0122} & \num{0.0158} & \num{0.0232} &
\num{0.0068} & \num{0.0207} & \num{0.0682} & \num{0.0269} & \num{0.0295} & \num{0.0373} \\
E5-base-V2 \cite{WangLiang2022TEbW} & &
\num{0.0075} & \num{0.0175} & \num{0.0380} & \num{0.0061} & \num{0.0095} & \num{0.0140} & \num{0.0041} & \num{0.0107} & \num{0.0317} & \num{0.0204} & \num{0.0203} & \num{0.0217} \\
GritLM-7B \cite{muennighoff2024generative} & &
\num{0.0240} & \num{0.0430} & \num{0.0965} & \num{0.0202} & \num{0.0265} & \num{0.0382} &
\num{0.0185} & \num{0.0375} & \num{0.0944} & \num{0.0606} & \num{0.0570} & \num{0.0614} \\
\midrule
\rowcolor[HTML]{F2F2F2}
\multicolumn{14}{l}{\bf Dense (multi-vector)} \\
GTE-ModernColBERT \cite{GTE-ModernColBERT} & &
\num{0.8850} & \num{0.9110} & \num{0.9295} & \num{0.8763} & \num{0.8856} & \num{0.8898} &
\num{0.4544} & \num{0.5824} & \num{0.7325} & \num{0.6102} & \num{0.6535} & \num{0.6847} \\
\midrule

\rowcolor[HTML]{F2F2F2}
\multicolumn{14}{l}{\bf Learned Sparse Retrieval (LSR)} \\
SPLADE \cite{lassance2024spladev3} & &
\num{0.4715} & \num{0.6645} & \num{0.8095} & \num{0.4058} & \num{0.4749} & \num{0.5078} &
\num{0.2399} & \num{0.4022} & \num{0.6135} & \num{0.3730} & \num{0.4277} & \num{0.4685} \\
\midrule

\rowcolor[HTML]{F2F2F2}
\multicolumn{14}{l}{\bf Enhanced Methods} \\
ReasonIR-8B \cite{shao2025reasonir} & &
\num{0.0200} & \num{0.0385} & \num{0.0885} & \num{0.0178} & \num{0.0243} & \num{0.0353} &
\num{0.0218} & \num{0.0580} & \num{0.1295} & \num{0.0905} & \num{0.0914} & \num{0.0922} \\
Set-Comp LSR (zero-shot) \cite{krasakis2025constructing} &  &
\num{0.4715} & \num{0.6645} & \num{0.8095} & \num{0.4058} & \num{0.4749} & \num{0.5078} &
\num{0.1986} & \num{0.2881} & \num{0.3677} & \num{0.2730} & \num{0.2579} & \num{0.2480} \\
LogiCol \cite{shen-etal-2025-logicol} &  &
\num{0.0030} & \num{0.0045} & \num{0.0150} & \num{0.0029} & \num{0.0033} & \num{0.0056} & \num{0.0026} & \num{0.0037} & \num{0.0168} & \num{0.0063} & \num{0.0067} & \num{0.0091} \\
Search-R1 (with BM25) \cite{jin2025search} &  &
\num{0.9385} & \num{0.9495} & \num{0.9600} & \num{0.9335} & \num{0.9376} & \num{0.9399} &
\num{0.5371} & \num{0.6583} & \num{0.8294} & \num{0.6510} & \num{0.6937} & \num{0.7295} \\
\bottomrule
\end{tabular}
\end{table*}

\section{Background} \label{sec:background}

\rqheading{Complex queries and set-compositional retrieval.}
Developing retrieval benchmarks with complex queries has gained significant attention in recent years; benchmarks such as BEIR~\cite{thakur2021beir} and MAIR~\cite{sun2024mair} evaluate retrieval models across heterogeneous tasks and instruction-following settings, yet their queries are predominantly topical: relevance is driven by semantic similarity between query and document content. Set-compositional queries pose a structurally different challenge. Here, the correct answer set is defined by the logical combination of multiple constraints (intersection, union, set difference), and a model must satisfy \emph{all} of them rather than match a single information need.
\textsc{QUEST}~\cite{malaviya-etal-2023-quest} formalizes this by constructing entity-seeking queries whose templates implicitly compose set operations over Wikipedia, with gold answer sets derived from Wikidata.
NevIR~\cite{weller-etal-2024-nevir} isolates the negation dimension, showing that even strong bi-encoders struggle to distinguish a query from its negated counterpart; a subsequent reproducibility study~\cite{10.1145/3726302.3730294} confirms these findings.
ExcluIR~\cite{zhang2025excluir} further demonstrates the difficulty of exclusionary queries for neural retrievers.

\rqheading{Limitations of neural IR under composition.}
Retrieval families differ in how they represent and match query--document pairs, which has direct implications for compositional queries.
Lexical methods (BM25) match on exact terms and are insensitive to implicit logical structure.
Dense single-vector bi-encoders~\cite{izacard2021contriever, WangLiang2022TEbW, WangLiang2023ITEw, muennighoff2024generative, qwen3embedding, weller2024promptriever} compress the full query into one embedding, making it difficult to enforce multiple independent constraints.
Late-interaction models~\cite{KhattabOmar2020CEaE, GTE-ModernColBERT} retain per-token representations and aggregate fine-grained similarities, potentially capturing constraint-level structure.
Learned sparse retrievers such as SPLADE~\cite{lassance2024spladev3} produce weighted term expansions that stay in the lexical space while learning relevance weighting.
LogiCol~\cite{shen-etal-2025-logicol} provides empirical evidence that standard contrastive training fails on conjunctive and negative constraints in \textsc{QUEST}, and proposes a modified contrastive objective to address this.
LIMIT~\cite{weller2025theoreticallimitationsembeddingbasedretrieval} takes a complementary angle: it constructs a synthetic corpus where relevance depends on explicit attribute predicates, showing that even \emph{atomic} queries can expose fundamental representational limits of single-vector embeddings when semantic overlap is removed.

\rqheading{Emerging reasoning-focused methods.}
Several recent methods explicitly target complex-query or reasoning-intensive retrieval, though they differ substantially in their inductive biases:
\emph{ReasonIR} \cite{shao2025reasonir} trains a dense retriever on synthetic reasoning tasks to improve retrieval on multi-step and compositional queries.
\emph{LogiCol}~\cite{shen-etal-2025-logicol} modifies the contrastive objective to be sensitive to set-theoretic structure in the query.
\emph{Set-Comp LSR}~\cite{krasakis2025constructing} composes learned sparse representations algebraically---performing vector addition and subtraction to mirror logical operators---without additional training.
\emph{Search-R1}~\cite{jin2025search} uses reinforcement learning to train an LLM to iteratively reformulate queries and aggregate search results.
\emph{OrLog}~\cite{hoveyda2026orlogresolvingcomplexqueries} and \emph{NS-IR}~\cite{xu-etal-2025-logical} adopt neuro-symbolic pipelines that decompose the query into logical sub-expressions before retrieval or re-ranking.
Comparing these methods requires a unified evaluation pipeline and shared benchmarks, because they optimize fundamentally different surrogates; synthetic data, modified contrastive losses, RL search policies, algebraic composition, or symbolic decomposition.

\rqheading{Research gap.}
Despite the breadth of prior work, two questions remained unanswered so far.
First, no existing study systematically compares all major retrieval families \emph{and} the emerging reasoning-focused methods under a single protocol for set-compositional queries.
Second, it is unclear to what extent reported gains reflect genuine constraint satisfaction rather than exploitation of semantic cues present in natural-language benchmarks.
Our study addresses both gaps: we benchmark all families on \textsc{QUEST} and \textsc{QUEST+Variants}, and introduce \textsc{LIMIT+} to disentangle semantic relatedness from predicate satisfaction, enabling controlled ablations by constraint type and compositional depth.

\section{Generating LIMIT+} \label{sec:limit+}

\subsubsection*{Motivation}
\textsc{QUEST} provides a realistic complex-query benchmark, but its queries are grounded in Wikipedia and therefore entangle set-compositional structure with rich lexical and world-knowledge cues.
To isolate the effect of \emph{compositional constraint structure} from semantic shortcuts, we derive \textsc{LIMIT+} from \textsc{LIMIT}~\cite{weller2025theoreticallimitationsembeddingbasedretrieval}, a synthetic benchmark whose corpus consists of entities described by explicit attribute lists (e.g., ``\textit{Olinda Posso likes Bagels, Hot Chocolate, Pumpkin Seeds, \ldots}''), so that relevance reduces to attribute predicate satisfaction rather than semantic relatedness.
Since \textsc{LIMIT} contains only \emph{atomic} queries, we extend it into a set-compositional benchmark by composing atomic predicates into \textsc{QUEST}-style logical templates.
For example, given the exclusion template $A \setminus B$, the query ``\textit{Who likes Thin Mints but not the Los Angeles Rams?}'' yields a gold set containing exactly those entities whose attribute list includes \textit{Thin Mints} and excludes \textit{the Los Angeles Rams}.

\subsubsection*{Generation procedure}
The \textsc{LIMIT} corpus contains 50{,}000 synthetic entities, each associated with a subset of 1{,}848 distinct attributes.
We parse each entity description to extract its attribute set $\mathcal{A}(e)$ and build an inverted index mapping each attribute to the set of entities possessing it. We instantiate seven query templates that mirror \textsc{QUEST}'s compositional structure, spanning three levels of depth (one to three predicates) and four logical operations:
\begin{itemize}[leftmargin=*, itemsep=1pt, topsep=2pt]
    \item \textbf{Atomic}: ``Who likes $A$?'' \hfill ($A$)
    \item \textbf{Disjunction}: ``$A$ or $B$''; ``$A$ or $B$ or $C$'' \hfill ($A \cup B$;\; $A \cup B \cup C$)
    \item \textbf{Conjunction}: ``$A$ and also $B$''; ``$A$ and also both $B$ and $C$'' \hfill ($A \cap B$;\; $A \cap B \cap C$)
    \item \textbf{Exclusion}: ``$A$ but not $B$''; ``$A$ and also $B$ but not $C$'' \hfill ($A \setminus B$;\; $(A \cap B) \setminus C$)
\end{itemize}
For atomic queries, we sample directly from \textsc{LIMIT}'s original query set.
For the remaining six templates, we sample one to three attributes and compute the gold answer set exactly via the corresponding set operation on the inverted index---e.g., for template $(A \cap B) \setminus C$, the gold set is $\{e \mid A \in \mathcal{A}(e) \wedge B \in \mathcal{A}(e) \wedge C \notin \mathcal{A}(e)\}$.
Uniqueness within each template is enforced by keying on the (sorted) attribute tuple.

\subsubsection*{Stratified sampling}
To prevent any single template from being dominated by queries with very large or very small answer sets, we apply stratified acceptance sampling over the gold-set size.
We define five buckets over the range $[1, 200]$ and aim for approximately equal counts per bucket within each template.
Queries with gold-set sizes outside $[1, 200]$ are discarded.
A two-phase procedure first fills each bucket to a per-bucket quota, then tops up with any remaining valid queries until the per-template limit is reached.

\subsubsection*{Statistics}
The final \textsc{LIMIT+} benchmark contains 700 queries (100 per template), with an overall mean gold-set size of ${\approx}$37 entities.
Mean gold-set sizes vary substantially across templates; from 1 (exclusion $A \setminus B$) to 128.3 (conjunction-with-exclusion $(A \cap B) \setminus C$), reflecting the inherent asymmetry of set operations over the attribute distribution.
Each query is stored with its natural-language form, gold entity identifiers, and metadata (template type, sampled attributes), enabling stratified evaluation by constraint type and compositional depth.
We release the generation script to enable exact reproduction and construction of further complex queries.

\section{Experimental Setup}
\label{sec:exp}

We evaluate a broad range of retrieval models spanning several paradigm families on multiple set-compositional benchmarks, considering six first stage rankers and four re-rankers.
The first-stage retrievers are evaluated in a zero-shot setting---no model is fine-tuned on the target datasets (except LogiCol, which is trained on \textsc{QUEST+Variants}).
Our experimental code is shared publicly\footnote{Repository \url{https://github.com/informagi/Complex-Set-Compositional-IR/}, last accessed April 28$^{th}$, 2026.}, and implementation details and prompts are provided in Appendix~\ref{app:experiments}.

\subsection{First-Stage Retrieval}

\subsubsection*{Lexical}
We use BM25~\cite{robertson1994okapi} as the lexical baseline, indexing all documents and retrieving the top-1{,}000 results per query.

\subsubsection*{Dense (single-vector)}
Single-vector bi-encoders encode query and document into one embedding each and score by dot product or cosine similarity.
We evaluate six models spanning two parameter scales:
Contriever~\cite{izacard2021contriever} and E5-base-V2~\cite{WangLiang2022TEbW} (both 110M parameters);
E5-Mistral-Instruct~\cite{WangLiang2023ITEw}, GritLM-7B~\cite{muennighoff2024generative}, and Promptriever~\cite{weller2024promptriever} (7B);
and Qwen3-Embedding~\cite{qwen3embedding} (8B). All models use (variants of) contrastive InfoNCE loss~\cite{ChenTing2020ASFf}, except for GritLM-7B, which uses language modeling together with contrastive embedding loss. 

\subsubsection*{Dense (multi-vector)}
Late-interaction models retain per-token embeddings and aggregate fine-grained token-level similarities at scoring time.
We evaluate GTE-ModernColBERT~\cite{GTE-ModernColBERT}, a ColBERT-style~\cite{KhattabOmar2020CEaE} retriever with 0.1B parameters. The model uses KL-Di\-ver\-gence loss and cross-entropy loss. 

\subsubsection*{Learned sparse retrieval (LSR)}
LSR models produce sparse, high-dimensional term-weight vectors that can be indexed with inverted indices while benefiting from learned relevance weighting.
We evaluate SPLADE-v3~\cite{lassance2024spladev3} (110M). Splade uses KL-Divergence loss and MarginMSE loss.

\subsubsection*{Enhanced methods}
We include four methods that explicitly target compositional or reasoning-intensive retrieval:
\begin{itemize}[leftmargin=*, itemsep=2pt, topsep=2pt]
    \item \textit{ReasonIR-8B}~\cite{shao2025reasonir} fine-tunes LLaMA-3.1-8B as a bi-encoder using public retrieval data augmented with synthetic reasoning-intensive queries. It uses InfoNCE contrastive loss.
    \item \textit{Set-Comp LSR}~\cite{krasakis2025constructing} constructs set-compositional query representations by performing linear algebra (addition, subtraction) on SPLADE vectors in a zero-shot setting, prior to training. Our experiments are limited to the zero-shot setting, as the trained checkpoint was not available. 
    \item \textit{LogiCol}~\cite{shen-etal-2025-logicol} fine-tunes multiple baselines, including E5-base-V2, with a modified contrastive objective that groups logically related queries in the same mini-batch. In our setting, we train E5-base-V2 on \textsc{QUEST+Variants} following the original setup in LogiCol. The reported results are all based on the trained checkpoint. 
    \item \textit{Search-R1}~\cite{jin2025search} trains a Qwen-2.5-7B model via reinforcement learning (proximal policy optimization (PPO) with exact-match reward) to iteratively reformulate queries and aggregate retrieved evidence. Our implementation follows a multi-run execution loop: we use the checkpoint trained on NQ and HotpotQA, with BM25 as the underlying retriever, and set the maximum number of turns to 5, retrieving the top-100 entities from which the top-3 are fed to the model. For evaluation, to be consistent with the rest of our setup, the aggregated top-100 BM25 results are used. The details of this implementation are in our repository.
\end{itemize}

\subsection{Re-ranking}
To isolate ranking capacity from candidate generation, we evaluate re-rankers under a curated candidate set.
For each query, the candidate pool consists of all gold documents plus five \emph{noise} documents (top-ranked non-gold BM25 hits, which are topically related but not relevant) and five \emph{irrelevant} documents (bottom-ranked BM25 hits).
This construction controls candidate-set difficulty while ensuring all gold documents are retrievable.

\subsubsection*{Cross-encoder}
MonoT5~\cite{NogueiraRodrigo2020DRwa} (220M) performs pointwise relevance scoring by feeding each query--document pair through a T5 encoder-decoder and computing the score as $p(\text{``true''})$.

\subsubsection*{LLM-based}
We evaluate two LLMs in a pointwise re-ranking setup using the fine-grained relevance scoring scheme of~\citet{zhuang2024beyond}: GPT-Oss~\cite{openai2025gptoss120bgptoss20bmodel} (120B) and Qwen-2.5-7B-Instruct~\cite{qwen2.5}.

\subsubsection*{Neuro-symbolic}
OrLog~\cite{hoveyda2026orlogresolvingcomplexqueries} decomposes complex queries into logical sub-expressions via an LLM parser (LLaMA-70B), estimates per-document predicate plausibilities with a smaller LLM (Qwen-2.5-7B), and aggregates them via probabilistic reasoning (ProbLog). We use the \textit{param+} configuration. OrLog adds no trainable parameters beyond its backbone LLMs.

\subsection{Datasets}

\subsubsection*{\textsc{QUEST} and \textsc{QUEST+Variants}.}
\textsc{QUEST}~\cite{malaviya-etal-2023-quest} contains 1{,}727 test queries over 325{,}427 Wikipedia entity documents, with queries generated from templates encoding up to two implicit set operations.\linebreak
\textsc{QUEST+Variants} \cite{shen-etal-2025-logicol} augments this to 11{,}854 queries over 361{,}646 documents by constructing companion queries that share the same atomic sub-queries but differ in logical connectives, enabling more comprehensive evaluation across operators.

\subsubsection*{\textsc{LIMIT} and \textsc{LIMIT+}}
\textsc{LIMIT}~\cite{weller2025theoreticallimitationsembeddingbasedretrieval} contains 1{,}000 atomic queries over a synthetic corpus of 50{,}000 entities, each described by an explicit attribute list.
\textsc{LIMIT+} (Section~\ref{sec:limit+}) extends this to 700 set-compositional queries (100 per template) over the same corpus, with gold sets computed exactly via set operations on the attribute inverted index.

\subsection{Evaluation Metrics}
For first-stage retrieval we report Recall@$k$ and nDCG@$k$ for $k \in \{5, 20, 100\}$, capturing both coverage and ranking quality at increasing depths.
For re-ranking we report nDCG@$\{5, 20\}$ and mean average precision (MAP) over the curated candidate set.

\begin{figure*}[htb]
  \centering
  \includegraphics[width=\textwidth]{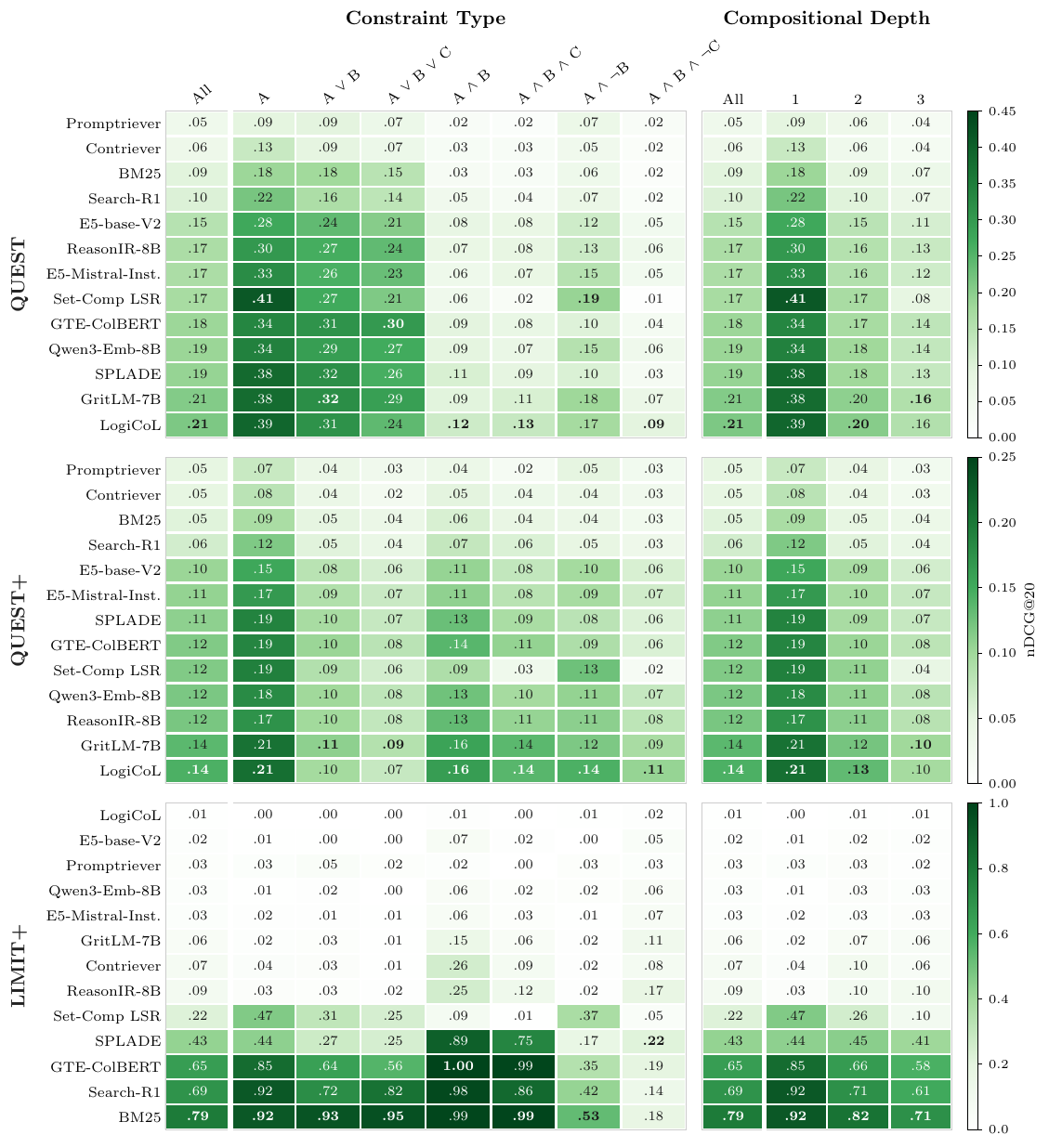}
  \caption{Constraint type and compositional depth analysis.}
  \label{fig:my-figure}
  \Description[Comparison]{Comparison across different Constraint Types and Compositional Depth for three datasets: QUEST, QUEST+ \& LIMIT+}
\end{figure*}

\section{Results}
\label{sec:res}
In this section we answer the raised research questions in the beginning of the paper. 

\rqheading{RQ1: How do major retrieval families and reasoning-targeted methods perform on set-compositional queries in QUEST and QUEST+?}
\noindent
Table~\ref{tab:repro_results_quest} reports first-stage retrieval results on \textsc{QUEST} and \textsc{QUEST+}.
\noindent
\paragraph{Retrieval paradigm-level separation}
On \textsc{QUEST}, lexical retrieval (BM25) is substantially outperformed by all neural methods (Recall@100 = 0.197).
Among neural retrievers, the best coverage is achieved by GritLM-7B and LogiCol, both attaining Recall@100 $\approx$ 0.42, followed by SPLADE (0.393) and Qwen3-Embedding (0.384).
LogiCol additionally achieves the highest early-cutoff recall (Recall@5 = 0.115, Recall@20 = 0.234).
In terms of ranking quality, GritLM-7B and LogiCol also lead on nDCG@100 (both 0.284), with SPLADE (0.262) and Qwen3-Embedding (0.257) following.
GTE-ModernColBERT, despite having only 0.1B parameters, achieves competitive nDCG (0.229 at cutoff 100) with lower recall (0.294), suggesting that late interaction ranks the items it retrieves effectively but surfaces fewer of them.

\paragraph{Generalization to \textsc{QUEST+}}
Moving from \textsc{QUEST} to \textsc{QUEST+}, absolute effectiveness drops substantially for all methods. Recall@100 roughly halves across the board, indicating that \textsc{QUEST+Variants} is a harder setting that exposes sensitivity to how constraints are expressed.
Despite this drop, the relative ordering among leading methods is largely preserved: LogiCol (Recall@100 = 0.217, nDCG@100 = 0.163) and GritLM (0.211, 0.156) remain the top performers, suggesting that the shift primarily affects absolute scores rather than fully reordering the approaches.

\paragraph{Reasoning-targeted methods}
Enhanced methods differ substantially in effectiveness.
LogiCol achieves the best or near-best scores across all cutoffs, competing with the strongest general-purpose retriever (GritLM-7B) despite being over 60$\times$ smaller (110M vs.\ 7B parameters).
ReasonIR-8B performs comparably to standard dense models of similar scale (Recall@100 = 0.364 on \textsc{QUEST}), offering no clear advantage for set-compositional queries despite its reasoning-oriented synthetic training data.
Search-R1 provides only marginal gains over BM25 (Recall@100 = 0.228 vs.\ 0.197); its RL-trained reformulation strategy, optimized for finding a single correct answer, appears poorly matched to the task of retrieving complete answer sets.
Set-Comp LSR, despite adding zero parameters, achieves Recall@100 = 0.289 and nDCG@100 = 0.217 by algebraically composing SPLADE vectors, though it falls short of SPLADE itself on coverage (0.289 vs.\ 0.393), indicating that naive vector arithmetic does not fully capture set-compositional semantics.

These results show that a \emph{reasoning-specific} training objective does not automatically translate into better set-compositional retrieval; what matters is the alignment between a model's inductive bias and the compositional structure of the task.

\rqheading{RQ2: Do gains observed on QUEST persist in a controlled setting in LIMIT+ where semantic similarity provides limited signal and relevance is driven by constraint satisfaction?}

Figure~3 reports first-stage retrieval results on \textsc{LIMIT} and \textsc{LIMIT+}.
The picture changes dramatically compared to \textsc{QUEST}.
On atomic \textsc{LIMIT} queries, BM25 (Recall@100 = 0.964), Search-R1 (0.960), and GTE-ModernColBERT (0.930) are near-perfect, while all dense single-vector models collapse below 0.15 and LogiCol falls to 0.015.
On the compositional \textsc{LIMIT+} queries, BM25 remains the strongest first-stage retriever (Recall@100 = 0.837), closely followed by Search-R1 (0.829) and GTE-ModernColBERT (0.733), while dense single-vector models stay below 0.13 and LogiCol remains near zero (0.017).
Methods that dominated on \textsc{QUEST} do not maintain their advantage once the opportunity to take `semantic shortcuts' is removed, confirming the observed gains on \textsc{QUEST} can be (partially) attributed to semantic relatedness rather than satisfying logical constraints.

\paragraph{Comparative analysis: Set-Comp LSR and LogiCol}

While both LogiCol~\cite{shen-etal-2025-logicol} and Set-Comp LSR~\cite{krasakis2025constructing} identify the inability of current retrievers to handle complex boolean queries, they approach the solution from a different angle, dictated by the underlying architecture. LogiCol focuses on dense retrievers, arguing that logical operators can be internalized through learning. In contrast, Set-Comp LSR relies on the lexical grounding in learned sparse retrievers, constructing compositional queries by performing linear algebra operations on the embeddings. 

In our benchmarking on QUEST (Figure~\ref{fig:my-figure}), LogiCol generally outperforms Set-Comp LSR with an overall nDCG of 0.21 compared to 0.17. This is most evident in cases of intersections such as \textit{A $\cap$ B} or \textit{A $\cap$ B $\cap$ C}, where LogiCol achieves scores of 0.12 and 0.13 compared to Set-Comp LSR's 0.06 and 0.02. However, for simple atomic queries A and single negations A $\cap \neg$ B, Set-Comp LSR demonstrates a slight advantage with 0.41 and 0.19 compared to LogiCol's 0.39 and 0.17. This shows that while LogiCol effectively teaches dense retrievers to handle complex queries, Set-Comp LSR remains on a competitive level using lexical grounding and algebraic manipulation of the embeddings. 

Nevertheless, we experience a performance shift going to LIMIT+ benchmarking. Here Set-Comp LSR proves significantly more robust with an overall nDCG of 0.22, while LogiCol achieves a score of 0.01. This contrast implies that LogiCoL's learned representations may be overfitted to the semantic distributions of the QUEST training data, leading to poor generalization when those semantic cues are removed. In contrast, Set-Comp LSR's zero-shot algebraic framework provides a more stable and generalizable baseline for logical evaluation.

\paragraph{Assessing template-specific gains in Set-Comp LSR}

The transition from context rich datasets like QUEST to constraint driven LIMIT+ dataset showcases differing level of robustness among sparse retrievers. We observe that adding logical context to the model via SetComp LSR only benefits specific templates such as A $\cap\neg$ B within the QUEST benchmark. In this scenario, SetComp LSR achieves an nDCG of 0.19, nearly doubling the 0.10 achieved by SPLADE. This suggests that the Set-Comp LSR zero shot framework effectively handles logical exclusion when semantic information is abundant. 

This performance increase is similarly reflected in the LIMIT+ benchmark, appearing not only in the A $\cap\neg$ B template but also in the A$\cup$ B template. In these specific logical scenarios, Set-Comp LSR achieves nDCG scores of 0.37 and 0.31, respectively. When compared against SPLADE, which only yields lower scores of 0.17 and 0.27, it may suggest that Set-Comp LSR improve template specific performance such as negation and union. However, this specialized robustness does not translate across all logical constraints from QUEST to LIMIT+. Specifically, for intersections A $\cap $ B and A $\cap $ B $\cap$ C, performance collapses to nDCG scores of 0.09 and 0.01, respectively. 

\paragraph{General picture}
Comparing the models across both benchmarks reveals a strong contrast in robustness against logical constraints. While the best performing models on QUEST, such as LogiCol (nDCG 0.21), GritLM-7B (nDCG 0.21) or Qwen3-Emb-8B (nDCG 0.19), establish a performance ceiling, they experience a near-total collapse when evaluated on LIMIT+ with nDCG varying from 0.01 to 0.06. This implies that these models rely heavily on semantic context rather than explicit logical constraints, the former of which is absent in the LIMIT+ dataset.

Apart from other models performing worse or better on the LIMIT+ data,  we observe an inverse effect in models such as BM25 and Search-R1, which demonstrate notable performance gains in this restricted setting. Specifically, BM25, whose performance on QUEST was limited to an nDCG of 0.09, now becomes the best performing model on the LIMIT+ data with an nDCG of 0.79. This is closely followed by Search-R1 who increased its nDCG from 0.10 on QUEST to 0.69 on LIMIT+.

\rqheading{RQ3: How does effectiveness change with increasing compositional depth and across constraint types?}

Figure~\ref{fig:my-figure} presents nDCG@20 stratified by constraint type (left column) and compositional depth (right column) across \textsc{QUEST}, \textsc{QUEST+Variants}, and \textsc{LIMIT+}.

\paragraph{Constraint type}
Across all three benchmarks, performance varies substantially by template.
On \textsc{QUEST}, atomic ($A$) and disjunctive queries ($A \cup B$, $A \cup B \cup C$) are comparatively easier for most models, while conjunction ($A \cap B$, $A \cap B \cap C$) and exclusion ($A \setminus B$, $A \cap B \setminus C$) templates yield markedly lower scores.
This pattern intensifies on \textsc{LIMIT+}: for conjunction templates, most dense models approach nDCG@20 of zero, whereas BM25, Search-R1, and GTE-ModernColBERT retain moderate effectiveness.
Exclusion templates reveal a different split: Set-Comp LSR shows a clear advantage for $A \setminus B$ (nDCG@20 $\approx$ 0.37 on \textsc{LIMIT+}, nearly doubling SPLADE's 0.17), confirming that algebraic vector subtraction captures negation more directly than learned representations.

\paragraph{Compositional depth}
Performance degrades monotonically as the number of atomic predicates increases from one to three.
On \textsc{QUEST}, the drop from depth one to depth three is moderate for leading models (e.g., GritLM nDCG@20 decreases from $\sim$0.35 to $\sim$0.13).
On \textsc{LIMIT+}, the degradation is steeper and more discriminating: BM25 and Search-R1 decline gradually, while dense single-vector models and LogiCol collapse to near-zero at depth two or three.
This confirms that compositional depth is a primary driver of difficulty, and that models which appear strong at depth one may fail entirely under deeper composition.

\paragraph{Takeaway}
Aggregate scores mask pronounced heterogeneity across constraint types and depths.
The stratified analysis reveals that no single model dominates across all templates: token-level methods (BM25, SPLADE, GTE-ModernColBERT) are more robust to increasing depth, algebraic composition (Set-Comp LSR) provides targeted gains on exclusion, and learned dense models degrade sharply once either conjunction or depth increases beyond one.

\sisetup{
  detect-weight=true,
  detect-family=true,
  table-number-alignment = center,
  round-mode=places,
  round-precision=3,
  table-format=1.3
}

\setlength{\tabcolsep}{3pt}

\begin{table}[ht]
\centering
\small
\renewcommand{\arraystretch}{1.08}
\caption{
Re-ranking results under a curated candidate set.
For each query, candidates are composed of all gold documents plus 5 \textit{noise} documents (top-ranked non-gold BM25 hits) and 5 \textit{irrelevant} documents (bottom-ranked BM25 hits).
Metrics show nDCG@5, nDCG@20, and MAP (computed per query and averaged). Scores in bold indicate the highest score in that column.
}
\label{tab:reranking_curated}
\resizebox{\columnwidth}{!}{%
\begin{tabular}{
  l l
  @{}r r r @{\hspace{\dsGutter}}
     r r r
}
\toprule
\multicolumn{2}{c}{} &
\multicolumn{3}{c}{\bf QUEST} &
\multicolumn{3}{c}{\bf LIMIT+} \\
\cmidrule(lr){3-5}\cmidrule(lr){6-8}

\multicolumn{2}{c}{} &
\multicolumn{2}{c}{\bf nDCG} & \multicolumn{1}{c}{\bf MAP} &
\multicolumn{2}{c}{\bf nDCG} & \multicolumn{1}{c}{\bf MAP} \\
\cmidrule(lr){3-4}\cmidrule(lr){5-5}
\cmidrule(lr){6-7}\cmidrule(lr){8-8}

\multicolumn{2}{c}{} &
\makecell{\scriptsize @5} & \makecell{\scriptsize @20} &
\makecell{\scriptsize (all)} &
\makecell{\scriptsize @5} & \makecell{\scriptsize @20} &
\makecell{\scriptsize (all)} \\
\midrule

\rowcolor[HTML]{F2F2F2}
\multicolumn{8}{l}{\bf Cross-encoder} \\
MonoT5 \cite{NogueiraRodrigo2020DRwa} & &
\num{0.155} & \num{0.530} & \num{0.402} &
\num{0.033} & \num{0.452} & \num{0.309} \\
\midrule

\rowcolor[HTML]{F2F2F2}
\multicolumn{8}{l}{\bf LLM-based} \\
GPT-Oss-120B \cite{openai2025gptoss120bgptoss20bmodel} & &
\bf\num{0.889} & \bf\num{0.915} & \bf\num{0.872} &
\bf\num{0.970} & \bf\num{0.967} & \bf\num{0.967} \\
Qwen-2.5-7B-Inst \cite{qwen2.5} & &
\num{0.751} & \num{0.821} & \num{0.736} &
\num{0.858} & \num{0.882} & \num{0.849} \\
\midrule

\rowcolor[HTML]{F2F2F2}
\multicolumn{8}{l}{\bf Neuro-symbolic} \\
OrLog (Qwen-2.5-7B-Inst) \cite{hoveyda2026orlogresolvingcomplexqueries} & &
\num{0.870} & \num{0.911} & \num{0.860} &
\num{0.920} & \num{0.927} & \num{0.917} \\
\bottomrule
\end{tabular}%
}
\end{table}

\rqheading{Re-ranking: LLMs, cross-encoder, neuro-symbolic}
To isolate ranking capacity from candidate generation, Table~\ref{tab:reranking_curated} evaluates re-rankers under a curated candidate set on \textsc{QUEST} and \textsc{LIMIT+}.

\paragraph{Cross-encoder}
MonoT5 fails on both benchmarks (nDCG@5 = 0.155 on \textsc{QUEST}, 0.033 on \textsc{LIMIT+}), indicating that a seq2seq cross-encoder trained on passage ranking lacks the capacity to handle multi-constraint relevance judgments.

\paragraph{LLM-based}
GPT-Oss achieves near-perfect re-ranking on both benchmarks (nDCG@5 = 0.889 on \textsc{QUEST}, 0.970 on \textsc{LIMIT+}; MAP = 0.872 and 0.967 respectively).
Qwen-2.5-7B-Instruct is also strong (nDCG@5 = 0.751 / 0.858), substantially outperforming MonoT5 despite using pointwise scoring.
The high performance on \textsc{LIMIT+}, where semantic cues are minimal, suggests that LLMs can reason about attribute constraints when the task is framed as classification rather than retrieval.

\paragraph{Neuro-symbolic}
OrLog achieves nDCG@5 = 0.870 on \textsc{QUEST} and 0.920 on \textsc{LIMIT+}, matching or exceeding GPT-Oss on \textsc{LIMIT+} while adding zero trainable parameters.
Its explicit logical decomposition and probabilistic aggregation appear well suited to set-compositional queries, and its strong \textsc{LIMIT+} performance confirms that the approach does not rely on semantic shortcuts.

\paragraph{Implications}
The contrast between first-stage retrieval (where most models struggle) and re-ranking (where LLMs and neuro-symbolic methods achieve nDCG@5 $>$ 0.87) localizes the primary bottleneck to candidate generation.
When gold documents are present in the candidate set, both LLMs and lightweight symbolic pipelines can resolve complex constraints effectively; the challenge is surfacing those candidates in the first place.

\begin{table}[t]
\centering
\caption{Reproduced vs.\ original recall scores (Ours/Orig). Q=QUEST, Q+=QUEST+Variants, L=LIMIT.}
\label{tab:repro_compare}
\small
\setlength{\tabcolsep}{4pt}
\renewcommand{\arraystretch}{1.05}

\begin{tabular}{@{}l l c c@{}}
\toprule
\textbf{Model} & \textbf{Dataset} & \textbf{R@20} & \textbf{R@100} \\
\midrule
BM25 \cite{shen-etal-2025-logicol,malaviya-etal-2023-quest} & Q & 0.104/0.104 & 0.197/0.197 \\
BM25 \cite{shen-etal-2025-logicol} & Q+ & 0.043/0.044 & 0.093/0.093 \\
Contriever \cite{shen-etal-2025-logicol} & Q & 0.079/0.064 & 0.146/0.136 \\
& Q+ & 0.032/0.032 & 0.073/0.073 \\
E5-base-V2 \cite{shen-etal-2025-logicol} & Q & 0.166/0.163 & 0.318/0.310 \\
& Q+ & 0.073/0.069 & 0.158/0.150 \\
LogiCol \cite{shen-etal-2025-logicol} & Q & 0.234/0.235 & 0.419/0.421 \\
& Q+ & 0.103/0.104 & 0.217/0.219 \\
\midrule
BM25 \cite{weller2025theoreticallimitationsembeddingbasedretrieval} & L & -- & 0.964/0.936 \\
GTE-ModernColBERT \cite{weller2025theoreticallimitationsembeddingbasedretrieval} & L & -- & 0.930/0.548 \\
Qwen3-Embedding \cite{weller2025theoreticallimitationsembeddingbasedretrieval} & L & -- & 0.048/0.048 \\
GritLM-7B \cite{weller2025theoreticallimitationsembeddingbasedretrieval} & L & -- & 0.097/0.129 \\
E5-Mistral-Instruct \cite{weller2025theoreticallimitationsembeddingbasedretrieval} & L & -- & 0.060/0.083 \\
\bottomrule
\end{tabular}
\end{table}

\section{Conclusion and Discussion}
\label{sec:conc}

We presented a reproducibility study of set-compositional information retrieval, benchmarking twelve retrieval models across four paradigm families and four reasoning-targeted methods on \textsc{QUEST}, \textsc{QUEST+}, and the newly introduced \textsc{LIMIT+} benchmark. The overarching finding is: \emph{absolute performance on set-compositional queries remains low across the board}. Even the best first-stage retrievers recover fewer than half the gold entities on \textsc{QUEST} (Recall@100 $\leq$ 0.42), and the strongest enhanced methods provide only incremental gains over well-tuned general-purpose models.
On \textsc{LIMIT+}, where semantic cues are removed, most dense models collapse to near-zero while lexical retrieval achieves Recall@100 $\approx$ 0.96, exposing that much of the progress observed on natural-language benchmarks reflects learned correlations with semantic context rather than genuine constraint satisfaction.
Three implications emerge from our results. Firstly, a step-by-step reasoning-specific training objective is insufficient for set-compositional retrieval.
LogiCol (110M) matches GritLM-7B on \textsc{QUEST} but collapses on \textsc{LIMIT+}; Search-R1's RL-trained reformulation barely improves over BM25; and the zero-shot algebraic approach of Set-Comp LSR generalizes more robustly than any learned method to the controlled setting.
This suggests that the field needs benchmarks that disentangle semantic relatedness from compositional structure, \textsc{LIMIT+} is one step in that direction. Secondly, the bottleneck lies in first-stage retrieval, not in ranking. Under controlled re-ranking, both LLMs (nDCG@5 up to 0.97) and a neuro-symbolic pipeline (OrLog; 0.92) achieve near-perfect constraint resolution, confirming that the challenge is surfacing the right candidates rather than judging them.
This points toward multi-stage pipelines that pair broad-coverage first-stage retrievers with constraint-aware re-rankers as a promising direction. Thirdly, aggregate metrics conceal critical failure modes. Our constraint-type and depth-stratified analyses show that conjunction and negation templates are markedly harder to satisfy than disjunction, and that performance degrades monotonically with compositional depth. These patterns are invisible in single-number comparisons and argue for fine-grained, template-level evaluation as a standard practice in complex-query IR.
\paragraph{Limitations.}
\textsc{LIMIT+} uses synthetic entities with explicit attribute lists, which simplifies the retrieval problem relative to natural text; its value lies in controlled diagnosis rather than as a standalone benchmark.
Our re-ranking evaluation uses a curated candidate set that guarantees gold-document presence, which upper-bounds real-world re-ranking performance.
We do have not evaluated generative retrieval or retrieval-augmented generation pipelines, which may handle compositional structure differently.

\paragraph{Future work.}
The gap between first-stage and re-ranking performance suggests that retrieve-then-reason architectures deserve systematic investigation for set-compositional queries.
Extending \textsc{LIMIT+} with deeper compositions (four or more predicates) and additional operators (e.g., symmetric difference, subset containment) would further stress-test model capabilities.
Finally, developing training objectives that explicitly encode set-theoretic structure---rather than relying on contrastive or RL surrogates---remains an open and pressing challenge.

\begin{acks}
This work has received funding from the Dutch Research Council (NWO) under project number NWA.1389.20.-183 and the European Union’s Horizon Europe research and innovation programme under grant agreement No 101070014
(OpenWebSearch.EU, \url{https://doi.org/10.3030/101070014}). 
All content represents the opinion of the authors, not necessarily shared or endorsed by their respective
employers and/or sponsors.
\end{acks}

\appendix
\section{Additional Experimental Details} \label{app:experiments}

\subsection{Prompts}
\subsubsection{ReasonIR-8B and GritLM-7B} 

We follow \cite{shao2025reasonir, muennighoff2024generative} query template:

\begin{promptbox}{Query Encoding Template}
{\ttfamily <|user|>\\
Given a query, retrieve relevant documents that answer the query\\
<|embed|>\\
\textit{\{query\_text\}}}
\end{promptbox}

\subsubsection{Reranking with Oss and Qwen}

For reranking with LLMs we use the following prompt template:

\begin{promptbox}{Prompting Template}
From a scale of 0 to 4, judge the relevance between the query and the document.  Return ONLY the integer score. \\ 
Query: \textit{\{query\_text\}}\\ 
Document: \textit{\{document\_text\}}\\ 
Output:
\end{promptbox}

\section{Reproducibility results}
In Table \ref{tab:repro_compare}, we compare our reproduced recall scores to those reported in prior work \cite{malaviya-etal-2023-quest,shen-etal-2025-logicol,weller2025theoreticallimitationsembeddingbasedretrieval}. For QUEST and QUEST+Variants, our reproduced  results closely match the original reports, suggesting that our preprocessing and evaluation pipeline is consistent for these datasets. However, we observe larger discrepancies on LIMIT, which is likely more sensitive to evaluation and retrieval details (e.g., dataset handling and scoring implementations).

\bibliographystyle{ACM-Reference-Format}

\bibliography{references}

\end{document}